\documentclass[final]{cvpr}

\usepackage{times}
\usepackage{epsfig}
\usepackage{graphicx}
\usepackage{amsmath}
\usepackage{amssymb}

\usepackage[pagebackref=true,breaklinks=true,colorlinks,bookmarks=false]{hyperref}

\def\cvprPaperID{2} 
\def\confYear{CVPR 2021}

\newcommand{\argmax}{\mathop{\mathrm{arg max}}}

\newcommand{\calB}{{\cal B}}

\newcommand{\calI}{{\cal I}}

\newcommand{\calO}{{\cal O}}
\newcommand{\calP}{{\cal P}}

\newcommand{\calT}{{\cal T}}

\newcommand{\vb}{{\bf b}}

\newcommand{\vg}{{\bf g}}

\newcommand{\vh}{{\bf h}}
\newcommand{\vp}{{\bf p}}

\newcommand{\vy}{{\bf y}}

\newcommand{\vA}{{\bf A}}

\newcommand{\vI}{{\bf I}}
\newcommand{\vM}{{\bf M}}

\newcommand{\vX}{{\bf X}}

\def\onedot{. }

\begin{document}
	
	\title{Know Your Surroundings: Panoramic Multi-Object Tracking by Multimodality Collaboration}
	
	\author{Yuhang He$^1$, 
		Wentao Yu$^2$, 
		Jie Han$^2$, 
		Xing Wei$^2$, 
		Xiaopeng Hong$^{3,4}$\thanks{Corresponding author.}, 
		Yihong Gong$^2$\\
		\textsuperscript{\rm 1}College of Artificial Intelligence , Xi'an Jiaotong University\\
		\textsuperscript{\rm 2}School of Software Engineering, Xi’an Jiaotong University \\
		\textsuperscript{\rm 3}School of Cyber Science and Engineering, Xi'an Jiaotong University \\
		\textsuperscript{\rm 4}Research Center for Artificial Intelligence, Peng Cheng Laboratory\\
		{\tt\small \{hyh1379478,yu1034397129,hanjie1997\}@stu.xjtu.edu.cn}, \tt\small \{weixing,hongxiaopeng,ygong\}@mail.xjtu.edu.cn
	}
	\maketitle

	\begin{abstract}
		In this paper, we focus on the multi-object tracking (MOT) problem of automatic driving and robot navigation. Most existing MOT methods track multiple objects using a singular RGB camera, which are prone to camera field-of-view and suffer tracking failures in complex scenarios due to background clutters and poor light conditions. To meet these challenges, we propose a MultiModality PAnoramic multi-object Tracking framework (MMPAT), which takes both 2D panorama images and 3D point clouds as input and then infers target trajectories using the multimodality data. The proposed method contains four major modules, a panorama image detection module, a multimodality data fusion module, a data association module and a trajectory inference model. We evaluate the proposed method on the JRDB dataset, where the MMPAT achieves the top performance in both the detection and tracking tasks and significantly outperforms state-of-the-art methods by a large margin (15.7 and 8.5 improvement in terms of AP and MOTA, respectively).
	\end{abstract}
	
	\section{Introduction}
	
	Multiple Object Tracking (MOT) aims to locate the positions of interested targets, maintains their identities across frames and infers a complete trajectory for each target. It has a wide range of applications in video surveillance~\cite{yang2016exploit,ren2018fusing}, custom behavior analysis~\cite{ferryman2009pets2009,he2020multi}, traffic flow monitoring~\cite{naphade20204th} and etc. Benefited from the rapid development of object detection techniques~\cite{ren2015faster,he2017mask,Cai_2019,Liu_2016,tian2019fcos,zhang2019freeanchor,law2018cornernet}, most state-of-the-art MOT trackers follow a \emph{tracking-by-detection} paradigm. They first detect targets in each image using modern object detectors and then associate these detection responses into trajectories by \emph{data association}. These methods have achieved significant improvement in recent years and became the main stream of MOT.
	
	\begin{figure}[t]
		\centering
		\begin{tabular}{l}
			\includegraphics[width=0.95\linewidth]{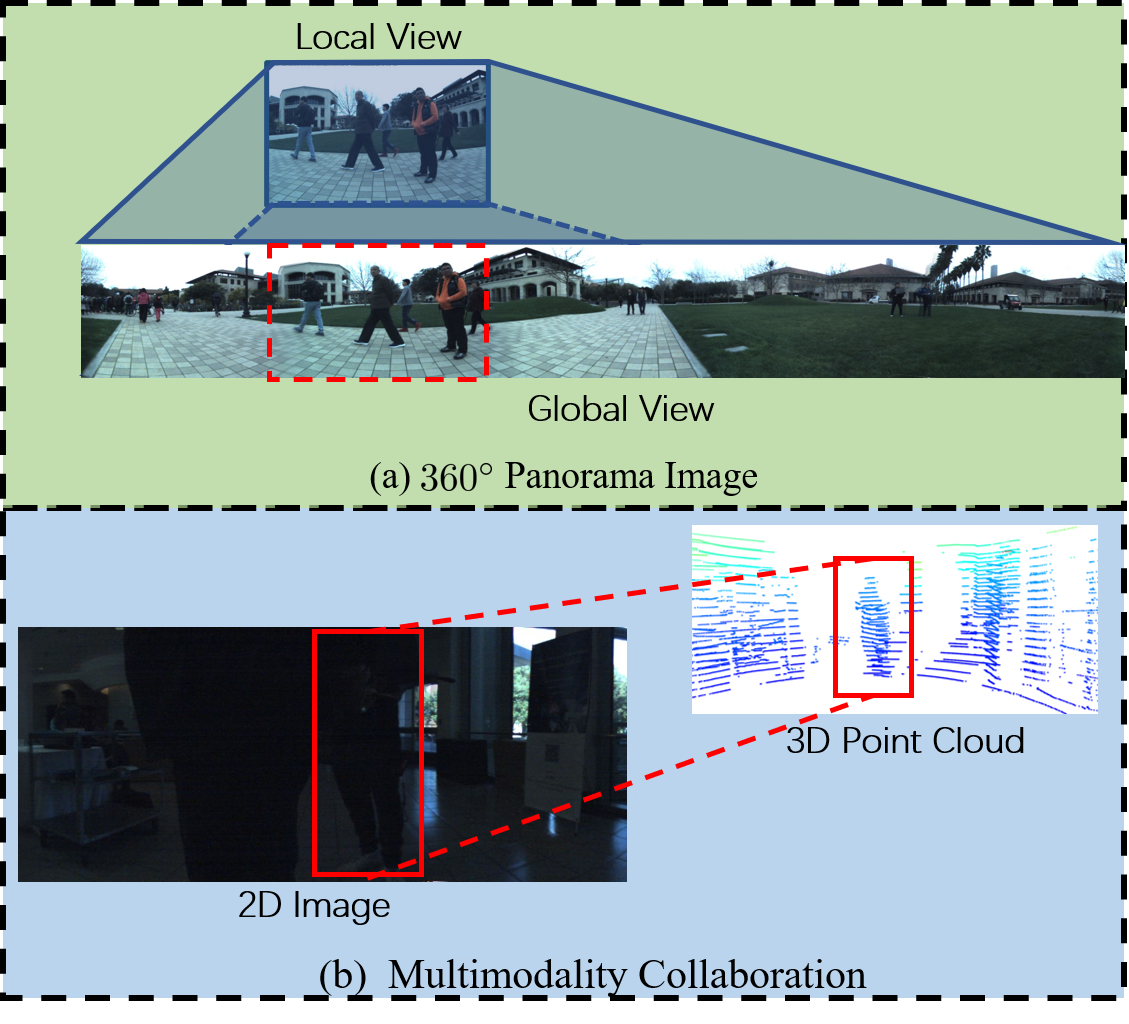}
		\end{tabular}
		\caption{Illustrations of multimodality panoramic multi-object tracking. (a) 360$^\circ$ panorama image. (b). Multimodality collaboration.}\label{fig:motivation}
	\end{figure}
	
\begin{figure*}[t]
		\centering
		\begin{tabular}{l}
			\includegraphics[width=0.95\linewidth]{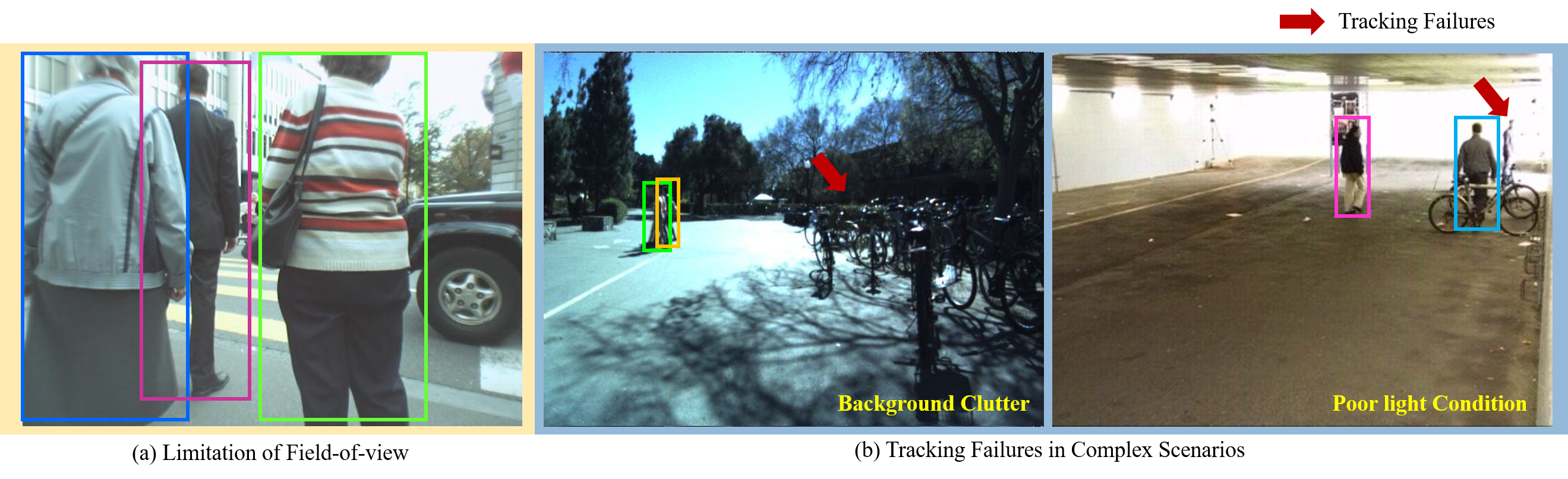}
		\end{tabular}
		\caption{Limitations of the single-modality single-camera tracking. (a) Limitation of camera field-of-view. (b) Tracking failures in background clutters and poor light conditions. The red arrows point to tracking failures.}\label{fig:MOT}
	\end{figure*}

	Accurate and efficient as they are, these methods are prone to camera field-of-view (FOV) and cannot handle the blind areas of camera views. Besides, limited to the properties of RGB cameras, these methods also have difficulties tracking targets in complex scenarios such as poor light conditions and background clutters. Figure~\ref{fig:MOT} illustrates a couple of tracking examples of the singular camera multi-object tracking. In (a), the MOT trackers track targets in a crowded scene. We can see that, the MOT trackers only generate sporadic trajectories while unconscious of the other targets in the surrounding. In (b), MOT trackers are failed to track targets due to background clutters and poor light conditions.
	These drawbacks of the single-camera MOT trackers prevent them from many important applications such as robot navigation~\cite{martin2019jrdb} and automatic driving~\cite{geiger2012we,caesar2020nuscenes}.
	
	To meet this challenge, we propose a MultiModality PAnoramic multi-object Tracking framework (MMPAT), which takes 2D 360$^\circ$ panorama images and 3D LiDAR point clouds as input and generates trajectories for targets by multimodality collaboration. The key insights of our MMPAT are twofold. First, \emph{a wider vision brings more information}. As shown in Figure~\ref{fig:motivation} (a), compared with the singular-camera MOT that tracks targets in a local view, taking the 360$^\circ$ panorama images as input enables us to have a global view of the surroundings and opens up opportunities for optimal tracking. Second, \emph{singular modality is biased while multimodality complement each other}. As shown in Figure~\ref{fig:motivation} (b), when the target in red bounding box is invisible due to poor light condition, the 3D point cloud supplements the target information for tracking. This provides a foundation for robust object tracking in complex scenarios. On this basis, we design the MMPAT algorithm with taking both the 2D 360$^\circ$ images and 3D point clouds as input. The proposed method is an online MOT method containing four major modules, 1) a panoramic object detection module, 2) a multimodality data fusion module, 3) a data association module and 4) a trajectory extension module. The Module 1 takes 2D images as input and outputs detection results for the panorama images. As panorama images are often long-width, the target responses in the feature maps of panorama images are narrow. This makes it difficult to locate the targets and generate accurate bounding boxes. To handle this problem, we design a split-detect-merge detection mechanism to detect targets in panorama images, which first splits panorama image into image slices, then detects targets in each slice, and finally merges detection responses from different slices. In Module 2, we fuse 2D images with 3D point clouds and append each detection with a 3D location characteristic. In Module 3, we match existing trajectories with newly obtained detections, where target appearance, motion and 3D location are exploited for data association. In Module 4, we generate accurate and complete trajectories for targets according to the data association results. The proposed MMPAT achieves the best performance in the detection and tracking tracks of the 2nd JRDB workshop\footnote{https://jrdb.stanford.edu/workshops/jrdb-cvpr21} and significantly outperforms state-of-the-art methods by a large margin (15.7 and 8.5 improvements on AP and MOTA, respectively).
	
	In summary, the contributions of this paper include:
	\begin{itemize}
		\item We propose a MultiModality PAnoramic multi-object Tracking framework (MMPAT) for robot navigation and automatic driving.
		\item We provide an efficient object detection mechanism to detect targets in panorama images.
		\item We design a 3D points collection algorithm to associate the point clouds with 2D images.
		\item The proposed method significantly improves the detection and tracking performance by a large margin.
	\end{itemize}
	
	\section{Related Work}
	\subsection{2D Multi-object Tracking}
	To date, most state-of-the-art MOT works follow the tracking-by-detection paradigm due to the rapid development of the detection techniques. According to whether frames following the current frame are available in the tracking process, tracking-by-detection paradigm based MOT methods can be divided into two subcategories: offline trackers and online trackers. 
	
	Offline MOT methods allow using (a batch of) the entire sequence to obtain the global optimal solution of the data association problem. A series of works~\cite{ma2018customized,sheng2018heterogeneous,tang2017multiple,wang2019exploit,yang2017hybrid,zamir2012gmcp} use graph models to link detections or tracklets (short trajectories) in the graph into trajectories. Ma~\emph{et al.}~\cite{ma2018customized} introduce a hierarchical correlation clustering (HCC) framework which builds different graph construction schemes at different levels to generate local, reliable tracklets as well as globally associated tracks. Wang~\emph{et al.}~\cite{wang2019exploit} utilize a graph model to generate tracklets by associating detections based on the appearance similarity and the spatial consistency measured by the multi-scale TrackletNet and cluster these tracklets to get global trajectories. A few methods~\cite{kim2015multiple,han2004algorithm,kim2018multi} tackle MOT problems by finding the most likely tracking proposals. Kim~\emph{et al.}~\cite{kim2015multiple} propose a novel multiple hypotheses tracking (MHT) method which enumerates multiple tracking hypothesis and selects the most likely proposals based on the features from long-term appearance models. There are also methods formulating the result optimization problem of MOT as a minimum cost lifted multicut problem~\cite{tang2017multiple}, a multidimensional assignment problem for multiple tracking hypothesis~\cite{kim2015multiple}, a Maximum Weighted Independent Set (MWIS) problem~\cite{sheng2018iterative} or a lifted disjoint paths problem~\cite{hornakova2020lifted}. Besides, a series of deep network based trackers are developed, such as Deep Tracklet Association (DTA)~\cite{zhang2020long}, bilinear LSTM (bLSTM)~\cite{kim2018multi}, Message Passing Netowrk (MPN)~\cite{braso2020learning}, and TrackletNet Tracker (TNT)~\cite{wang2019exploit}. There are other methods improve the performance of MOT by Tracklet-Plane Matching (TPM)~\cite{peng2020tpm} and Correlation Co-Clustering (CCC)~\cite{keuper2018motion}.
	
	Online MOT methods require that only the information in the current frame and the previous frame can be used to predict the tracking result of current frame, and the tracking result of the previous frame cannot be modified based on the information of the current frame. A large number of research studies~\cite{wojke2017simple,chu2019online,xiang2015learning,xu2019spatial,sadeghian2017tracking} utilize bipartite matching to tackle online MOT problems. Wojke~\emph{et al.}~\cite{wojke2017simple} divide the existing trajectories and new detections into two disjoint sets, and tackle the trajectory-detection matching problems by the Hungarian algorithm~\cite{munkres1957algorithms}. The method~\cite{sadeghian2017tracking} uses a Recurrent Neural Networks (RNN) to integrate Appearance, Motion and Interaction information (AMIR) to jointly learn robust target representations. A series of deep learning approaches are proposed to measure the similarity between a target and a tracklet, like Spatial Temporal Attention Mechanism (STAM)~\cite{chu2017online}, Recurrent Autoregressive Network (RAN)~\cite{fang2018recurrent}, Dual Matching Attention Network (DMAN)~\cite{zhu2018online}, and Spatial-Temporal Relation Network (STRN)~\cite{xu2019spatial}. In FAMNet~\cite{chu2019famnet}, Feature extraction, Affinity estimation and Multi-dimensional assignment are integrated into a single Network. Besides, there are several works that incorporate the technologies from other fields, such as Tracktor++~\cite{bergmann2019tracking} leverages bounding box regression from object detection, Instance Aware Tracking (IAT)~\cite{chu2019online} leverages the idea of model updatng from single object tracking, and GSM~\cite{liugsm} leverages the graph matching module from target relations.

	\subsection{3D Multi-object Tracking}
	
	\begin{figure*}[t]
		\centering
		\begin{tabular}{l}
			\includegraphics[width=0.95\linewidth]{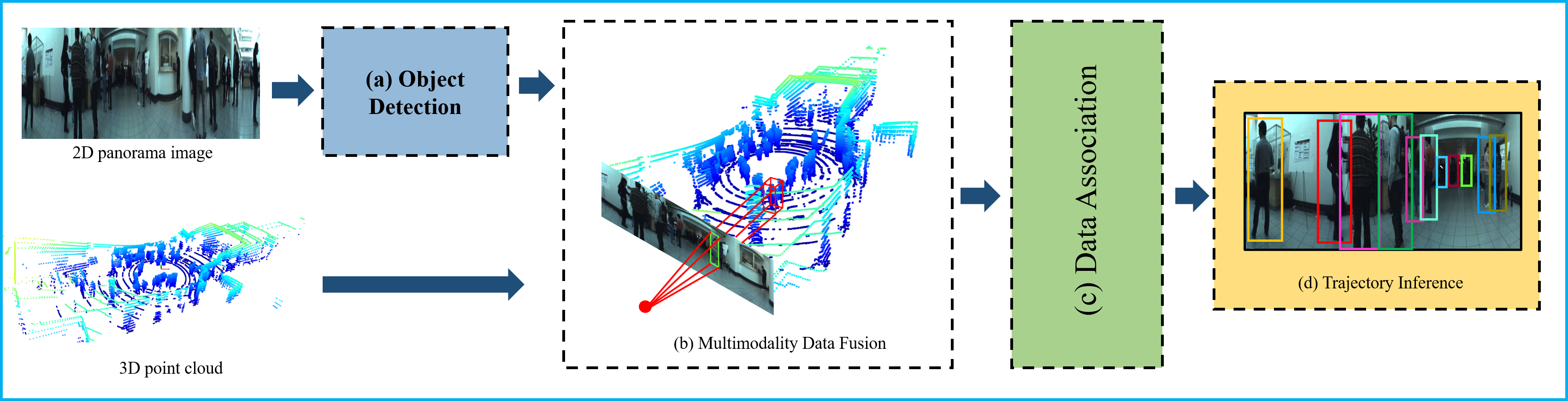}
		\end{tabular}
		\caption{Overview of the MMPAT framework. The inputs are 2D image frames and 3D point clouds. The proposed method contains four major modules. a) The object detection module. b) The Multimodality data fusion module. c) The data association module. d) The Trajectory inference module.}\label{fig:framerwork}
	\end{figure*}
	
	~\textbf{3D Object Detection.} There is a large literature on the use of instant sensor data to detect 3D object in the domain of autonomous driving. Depending on the modality of input data, 3D object detectors can be roughly divided into three categories: monocular image-based methods, stereo imagery-based methods, and LiDAR-based methods. Given a monocular image, early 3D object detection works~\cite{zeeshan2014cars,chen2016monocular,chabot2017deep,kundu20183d} usually exploit the rich detail information of the 3D scene representation to strengthen the understanding of 3D targets, like semantic and object instance segmentation, shape features and location priors, key-point, and instance model, while later state-of-the-art studies~\cite{xu2018multi,ku2019monocular,qin2019monogrnet,ma2019accurate,brazil2019m3d} pay more attention to 3D contexts and the depth information encoding from multiple levels for accurate 3D localization. Compared with the monocular-image based methods, stereo-imagery based methods~\cite{chen20173d,li2019stereo,wang2019pseudo,pon2020object} add additional images with known extrinsic configuration and achieve much better 3D object detection accuracy.The method~\cite{chen20173d} first generates high-quality 3D object proposals with stereo imagery by encoding depth informed features that reason about free space, point cloud densities and distance to the ground, and employs a CNN on these proposals to perform 3D object detection. Stereo R-CNN~\cite{li2019stereo} exploits object-level disparity information and geometric-constraints to get object detection by stereo imagery alignment. Wang~\emph{et al.}~\cite{wang2019pseudo} convert image-based depth maps generated from stereo imagery to pseudo-LiDAR representations and apply existing LiDAR-based detection approaches to detect object in 3D space. In addition to image-based methods, there is abundant literature~\cite{li2016vehicle,engelcke2017vote3deep,zhou2018voxelnet,lang2019pointpillars,qi2018frustum,shi2019pointrcnn} directly using the 3D information from the LiDAR point cloud to detect 3D objects. Several works~\cite{engelcke2017vote3deep,zhou2018voxelnet,lang2019pointpillars} sample the unstructured point cloud as a structured voxel representation and encode the features using 2D or 3D convolution networks to detect object. Methods~\cite{li2016vehicle,yang2018pixor} utilize conventional 2D convolutional networks to achieve 3D object detection by projecting point clouds to the front or bird's-eye views. Besides, there are also methods~\cite{qi2018frustum,shi2019pointrcnn} directly employ raw unstructured point clouds to localize 3D objects with the help of PointNet~\cite{qi2017pointnet} encoder. Moreover, there are other methods~\cite{chen2017multi,liang2018deep,ku2018joint,liang2019multi} fusing LIDAR point clouds and RGB images at the feature level for multi-modality detection.
	
	~\textbf{3D Object Tracking.} Due to the success of the tracking-by-detection paradigm in 2D object tracking, many 3D object tracking methods also follow this paradigm. Based on the 3D detection results, methods~\cite{osep2017combined,scheidegger2018mono,simon2019complexer} utilize filter-based (Kalman filter; Poisson multi-Bernoulli mixture filter) algorithm to continuously track 3D objects, while Hu~\emph{et al.}~\cite{hu2019joint} design an LSTM-based module using data-driving approaches to directly learn the object motion for more accurate long-term tracking. However, the loss of information caused by decoupling detection and tracking may lead to sub-optimal solutions. Benefit from stereo images, the method~\cite{engelmann2017samp} focuses on reconstructing the object using 3D shape and motion priors, and the method~\cite{li2018stereo} exploits a dynamic object bundle adjustment (BA) approach which fuses temporal sparse feature correspondences and the semantic 3D measurement model to continuously track the object, while the performance on 3D localization for occluded objects is limited. From another aspect, Luo~\emph{et al.}~\cite{luo2018fast} encode 3D point clouds into 3D voxel representations and jointly reason about 3D detection, tracking and motion forecasting so that it is more robust to occlusion as well as sparse data at range.

	\section{Methodology}
	In this section, we first overview the framework of our proposed method and then provide detailed descriptions of the key techniques.
	
	\subsection{Framework Overview}
	As illustrated in Figure~\ref{fig:framerwork}, the proposed MMPAT is an online MOT method containing four major modules: 1) an object detection module to locate targets in the panorama images, 2) a multimodality data fusion module to associate 3D point clouds with 2D images, 3) a data association module to match existing trajectories with newly obtained detections and 4) a trajectory inference module to generate trajectories for targets. In the following, we provide detailed descriptions of each module.

	\subsection{Object Detection in Panorama Image}
	Compared with object detection on ordinary-size images (such as 720P and 1080P images), there are two additional challenges that need to be solved with the panorama images. First, most two-stage object detectors resize the input images into a fixed size and then generate region proposals on the feature maps. However, as the 2D panorama images have a long width, the target responses are narrow and feeble in the feature maps of resized images. This makes it difficult to locate the target in the feature maps and generate accurate proposals for the targets. Second, in panorama images, the size of targets often varies in a large range due to perspective changes. This leaves a challenging problem to handle the size variations of targets for accurate object detection. To tackle these problems, we design an object detection algorithm for panorama image. As shown in Figure~\ref{fig:det}, we first split the panorama image into several image slices along the image width. Then, we detect objects in each image slice following a cascade detection paradigm~\cite{cai2018cascade}. In the end, detections responses from different image slices are merged together using NMS~\cite{bodla2017soft}.  
	
	\subsubsection{Detection Pipeline}
	\textbf{Panorama image split}. Given the panorama image $\vI_t$ at time $t$, we first split the image $\vI_t$ into $N$ image slices $\calI_t=\{\vI_t^n\}_{n=1}^N$, where the image slices $\calI_t$ are obtained by splitting image $\vI_t$ along the width dimension with an overlap of 0.2. 
	
\begin{figure}[t]
		\centering
		\begin{tabular}{l}
			\includegraphics[width=1.0\linewidth]{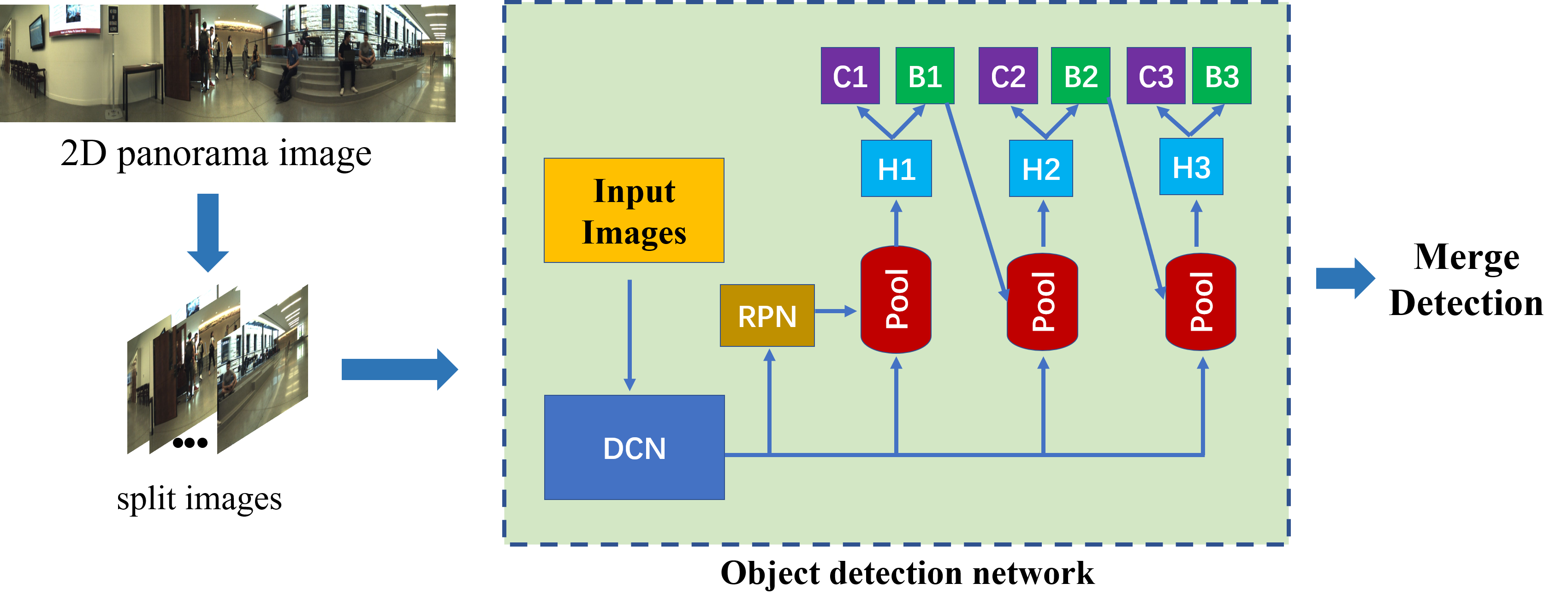}
		\end{tabular}
		\caption{Pipeline of object detection. The ``DCN" denotes deformable convolution network, ``RPN'' denotes region proposal network, ``pool'' denotes pooling layers, ``H'' denotes regression head layer, ``B'' and ``C'' denote bounding box and classification, respectively.}\label{fig:det}
	\end{figure}

	\textbf{Cascade object detector}. We then detect objects in each image slice $\vI_t^n$ using a cascade object detector. As shown in Figure~\ref{fig:det}, the object detector is composed of three components, \emph{i.e.}, a deformable convolution network, a region proposal network and a cascade detection header. In the deformable convolution network, we take the ResNet50 architecture~\cite{he2016deep} as our backbone, with the fully-connected layers and last pooling layer removed. To handle the target size variations, a deformable convolution~\cite{dai2017deformable} is employed, which adds 2D offsets to the sampling location of standard convolutions and enables free form deformations of the sampling grid. Then, taking the feature maps, a region proposal network~\cite{ren2015faster} is adopted to generate proposals of targets. Taking the region proposals and feature maps as input, the cascade detection header iteratively regresses the bounding boxes to produce more accurate bounding boxes. At each regression layer $l$, there is a classifier $h_l$ and a bounding box regressor $f_l$. For a bounding box $\vb$, the cascade object detector iteratively regresses the bounding box, which can be written as:
	\begin{equation}
	f(x,\vb)=f_L\circ f_{L-1}\circ \dots f_1(x,\vb),
	\end{equation}
	where $x$ is the input feature map and $L$ is the total number of regression layers. At the inference, the regression takes the region proposals as input and iteratively regresses bounding boxes. We use $\vb_0$ to denote the input region proposals and $\vb_l=f_{l}(x_{l},\vb_{l-1})$ to denote the output bounding box of the $l$-$th$ regressor. 
	
	\textbf{Detection merge}. We use $\calB_t(i)$ to denote the bounding box collection of the $i$-$th$ image slice $\vI_t^i$. We merge detection responses from all the image slices by Non-Maximum Suppression (NMS)~\cite{bodla2017soft}: 
	\begin{equation}
	\calB_t=\text{NMS}(\calB_t(1),...,\calB_t(N)),
	\end{equation}
	where $\calB_t$ denotes the detection set of panorama image $\vI_t$. We use $\calB_t^i$ to denote the $i$-$th$ detection in $\calB_t^i$.

	\subsubsection{Loss Function}
	For each regression layer $l$ of the cascade object detector, the loss function is composed of two parts: bounding box regression and classification.
	
	\textbf{Bounding box regression}. The objective of bounding box regression is to refine a candidate bounding box $\vb_l=(\vb_l^x,\vb_l^y,\vb_l^w,\vb_l^h)$ into a ground-truth bounding box $\vg_l=(\vg_l^x,\vg_l^y,\vg_l^w,\vg_l^h)$, where $(*^x,*^y)$ are the coordinate of bounding box center and $*^w$ and $*^h$ are the width and height, respectively. Transforming this objective into loss function, we have:
	\begin{equation}
	\text{L}_{loc}(\vb_l,\vg_l) = \sum_{j\in \{x,y,w,h\}} \text{smooth}_{L1}(\vb_l^j-\vg_l^j),
	\end{equation}
	in which 
	\begin{equation}
	\text{smooth}_{L1}(a) = \left\{ 
	\begin{array}{ll}
	0.5~a^2, &\text{if}~|a|<1, \\
	|a|-0.5, &\text{else}.
	\end{array}
	\right.
	\end{equation}

	\textbf{Classification}. We adopt the cross-entropy loss to optimize the classification header. We use $\vy_l$ to denote the one-hot ground-truth label of $\vb_l$ and use $\vp_l=h_l(\vb_l)$ to denote the output classification vector of $\vb_l$. Then the classification loss function can be written as:
	\begin{equation}
	L_{cls}(\vb_l,\vy_l) = \text{CrossEntropy}(\vp_l,\vy_l),
	\end{equation}
	in which
	\begin{equation}
	\text{CrossEntropy}(p,q) = -\sum_{i}^{C+1} p(i)\cdot log(q(i)),
	\end{equation}
	where $C$ is the number of classes and $p(i)$ (or $q(i)$) is the $i$-$th$ element of the vector $p$ (or $q$).
	On this basis, the loss function of the cascade object detector can be formulated as:
	\begin{equation}
	L_{total} (\vb_l,\vg_l,\vy_l) = \sum_{l=1}^L  L_{cls}(\vb_l,\vy_l)+\delta(\vy_l)\cdot L_{loc}(\vb_l,\vg_l),
	\end{equation}
	where $\delta(\vy_l)=0$ if $\vy_l$ belongs to background class and $\delta(\vy_l)=1$ for otherwise.

	\subsection{Multimodality Data Fusion}
\begin{figure}[tb]
		\centering
		\begin{tabular}{l}
			\includegraphics[width=0.85\linewidth]{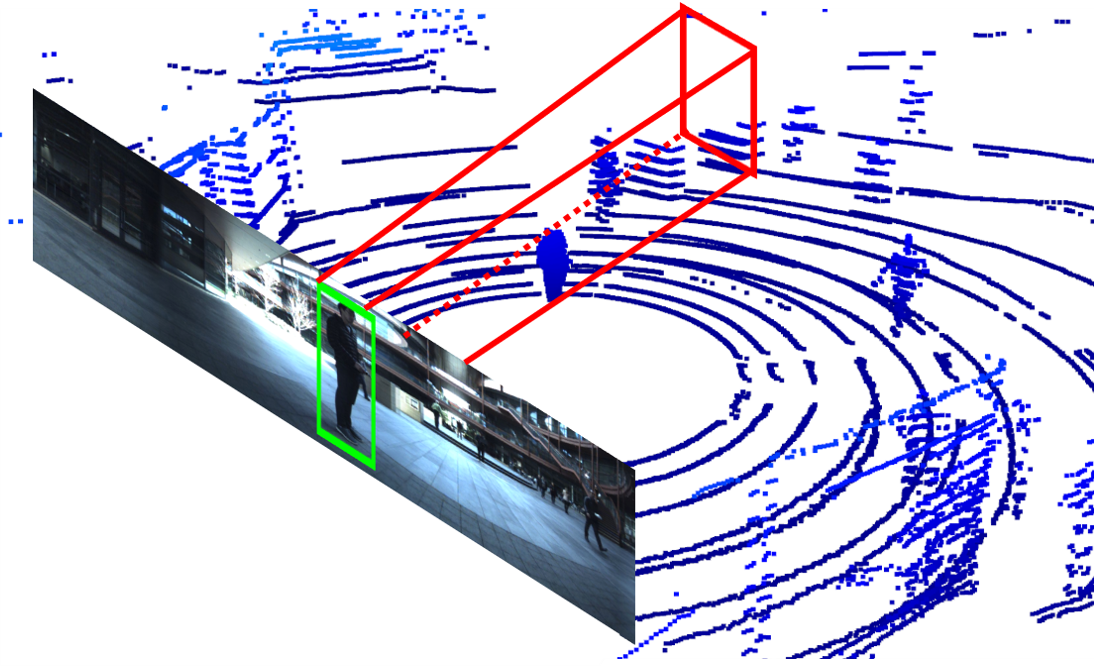}
		\end{tabular}
		\caption{Illustration of multimodality data fusion.}\label{fig:frustum}
	\end{figure}

	As shown in Figure~\ref{fig:frustum}, this module aims to associate detections with 3D points and append each detection $\calB_t^i$ with a 3D location characteristic $l_t^i$. The collection of 3D points for a detection contains two steps. First, we perform instance segment in the 2D bounding box to filter out the background clutters. Then, we collect 3D points of the target based on 3D-to-2D projection. 
	
	Specifically, let $\vM$ be the projection matrix from 3D point cloud to 2D image plane, $\Omega_{box}$ be the collection of foreground pixels of the 2D bounding box and $\Omega_{ptc}$ be the collection of 3D points in the point cloud. We collect 3D points of the target by:
	\begin{equation}
	\calP = \{\vh~|~\forall~\vh\in\Omega_{ptc}, \text{if}~\rho(\vh;\vM)\in \Omega_{box}\},
	\end{equation}
	where $\rho(\vh;\vM)$ projects the input 3D point $\vh$ to 2D pixel using the input projection matrix $\vM$. For computation efficiency, similar to~\cite{qi2018frustum}, we define a 3D frustum search space according to the 2D bounding box and then project 3D points to image plane within the search space. We use $\calP_t^v$ to denote the 3D points of detection $\calB_t^v$, the 3D location $l_t^v$ of $\calB_t^v$ is obtained by averaging the points in $\calP_t^v$, \emph{i.e.}, $l_t^v=\text{average}(\calP_t^v)$.

	\subsection{Data Association}
	We use $\calT_{t-1}=\{\calT_{t-1}^1,...,\calT_{t-1}^{K_{t-1}}\}$ to denote the collection of trajectories at time $t-1$, where $K_{t-1}$ is the number of trajectories. Each trajectory $\calT_{t-1}^i$ is made of a serious tuples:
	\begin{equation}
	\calT_{t-1}^i=\{(a_k^i,b_k^i,l_k^i), k\in \varphi_{t-1}^i\},
	\end{equation}
	where $\varphi_{t-1}^i$ is the time index set of the trajectory $\calT_{t-1}^i$, $a_k^i$, $b_k^i$, $l_k^i$ are the appearance feature, bounding box and location of the target at time $k$. Taking the bounding box collection $\calB_t=\{\calB_t^1,...,\calB_t^{Q_t}\}$, where $Q_t$ is the number of detections at time. The objective of data association is to match the newly obtained bounding boxes $\calB_{t-1}$  with the existing trajectories $\calT_{t-1}$, and manage target trajectories according to the matching result.
	
	We formulate the data association as a bipartite graph matching problem, where we first compute a pairwise trajectory-detection affinity matrix between trajectoreis and detections and then solve the matching problem using the Hungarian algorithm~\cite{kuhn1955hungarian}. 
	
	\subsubsection{Affinity Measurement}
	We use $\vA\in \mathbb{R}_{\geq 0}^{K_{t-1}\times Q_{t}}$ to denote the pairwise affinity matrix of $\calT_{t-1}$ and $\calB_t$, where each element $\vA(u,v)$ in $\vA$ denotes the affinity between $\calT_{t-1}^u$ and $\calB_t^v$. The larger $\vA(u,v)$ is, the higher affinity of $\calT_{t-1}^u$ and $\calB_t^v$ is. We compute the affinity $\vA(u,v)$ score of each trajectory-detection pair using the appearance, motion and 3D location information:
	\begin{equation}
	\vA(u,v) = \psi_{app}(\calT_{t-1}^u,\calB_t^v)+\psi_{mot}(\calT_{t-1}^u,\calB_t^v)+\psi_{loc}(\calT_{t-1}^u,\calB_t^v),
	\end{equation}
	where $\psi_{app}(\cdot,\cdot)$, $\psi_{mot}(\cdot,\cdot)$ and $\psi_{loc}(\cdot,\cdot)$ compute the appearance, motion and location affinity of the input trajectory and detection, respectively.
	
	\textbf{Appearance similarity}. The appearance similarity is computed by an averaged cross-correlation between the trajectory and detection appearance features, which can be written as: 
	\begin{equation}
	{\varphi_{app}(\calT_{t-1}^u,\calB_t^v)}=\frac{\sum_{\forall k\in \tau_{t-1}^u}\big[e^{k-t}\cdot \gamma (a_k^u,\phi(\calB_t^v))\big]}{\sum_{\forall k\in \tau_{t-1}^u}e^{k-t}}, \label{EQ:app_sim} 
	\end{equation}
	where $\tau_{t-1}^u$ is the collection of time index of trajectory $\calT_{t-1}^u$, $\phi(\cdot)$ is a feature extractor and $\gamma(\cdot,\cdot)$ outputs the cross-correlation score of the input features.  
	
	\textbf{Motion affinity}.
	The motion affinity is calculated by computing the Intersection-over-Union (IoU) between a predicted bounding box $\calO_t^i$ and detection $\calB_t^v$:
	\begin{eqnarray}
	{\varphi_{mot}(\calT_{t-1}^u,\calB_t^v)} = {\text{area}(\calO_t^u\cap \calB_t^v)}/{\text{area}(\calO_t^u\cup \calB_t^v)},
	\end{eqnarray}
	where {$\calO_t^u=\Phi(\calT_{t-1}^u)$} is a predicted bounding box according to the input trajectory $\calT_{t-1}^u$ using the Kalman filter~\cite{kalman1960new}.
	
	\textbf{Location proximity}. We calculate the 3D location proximity between a trajectory $\calT_{t-1}^u$ and a detection $\calB_t^v$ by:
	\begin{equation}
	{\varphi_{loc}(\calT_{t-1}^u,\calB_t^v)} = \sum_{k\in \tau_k^u} \frac{\sigma_{t}(k,t)\cdot \sigma_{l}(\calT_{t-1}^u(k)_{loc},l_t^v)}{|\tau_k^u|},
	\end{equation}
	where $\tau_{t-1}^u$ is the time index set of trajectory $\calT_{t-1}^u$, $\calT_{t-1}^u(k)_{loc}$ is the 3D location of trajectory $\calT_{t-1}^u$ at time $k$. The $\sigma_{t}(\cdot,\cdot)$ and $\sigma_{l}(\cdot,\cdot)$ output normalized time distance and location distance using two RBF kernels, respectively.
	
	\subsubsection{Bipartite Graph Matching}
	Given the trajectory-detection affinity matrix $\vA$, we aim to calculate a matching matrix $\vX_t\in \{0,1\}^{K_{t-1}\times Q_t}$ according to $\vA$. Each element $\vX(u,v)$ in $\vX$ corresponds {to} the matching (\emph{i.e.}, \vX(u,v)=1) and non-matching (\emph{i.e.}, \vX(u,v)=0) between trajectory $\calT_{t-1}^u$ and detection $\calB_t^v$.  The bipartite graph matching can be solved by the following optimization problem:
	\begin{eqnarray}
	\begin{split}
	& \vX^*=\argmax_{\vX} ~||\vA\odot \vX||_2,\\
	& ~~~s.t. ~~~~\forall~u,~~\sum \vX(u,:) \leq 1, \\
	& ~~~~~~~~~~~~~\forall~v,~~\sum \vX(:,v) \leq 1,
	\end{split}
	\end{eqnarray}
	where $\odot$ denotes element-wise matrix multiplication and $||\cdot||_2$ outputs the L2-norm of input matrix. The constraints ensure the mutual exclusion of trajectories, where each detection will be occupied with at most one trajectory. On this basis, the optimization problem can be efficiently solved by the Hungarian algorithm~\cite{kuhn1955hungarian}.
	
	\subsection{Trajectory Inference}
	\begin{figure}[t]
		\centering
		\begin{tabular}{l}
			\includegraphics[width=0.9\linewidth]{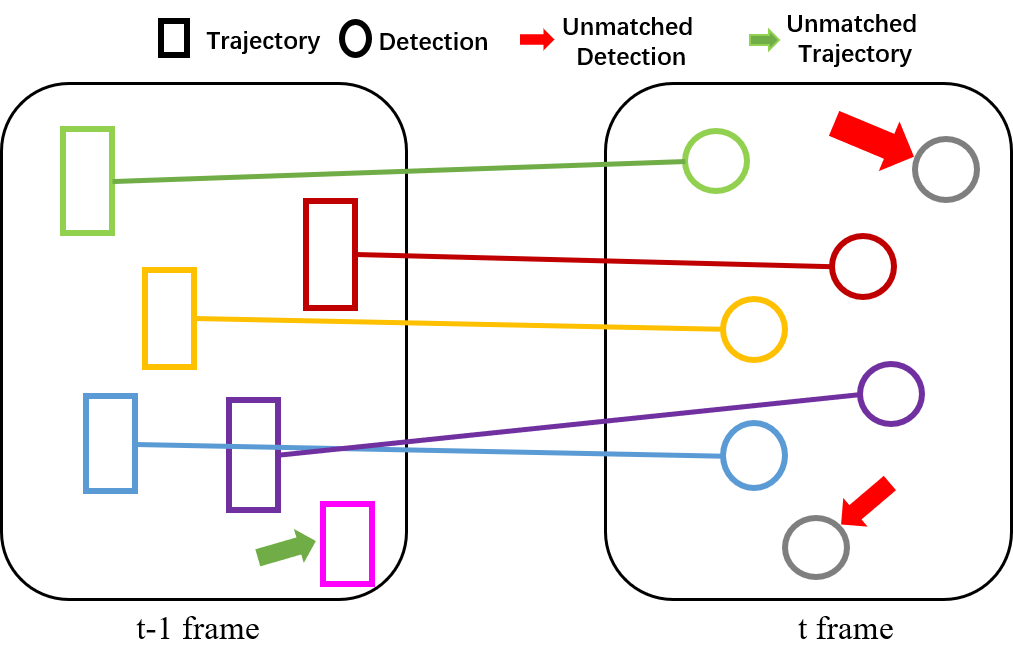}
		\end{tabular}
		\caption{Illustration of data association. The rectangles and circles denote trajectories and detections, respectively. The lines indicate the linkage between trajectories and detections.}\label{fig:data_association}
	\end{figure}
	
	In this module, we manage the target trajectories according to the matching matrix $\vX$. As shown in Figure~\ref{fig:data_association}, there are three conditions of the matching results: a) detection $\calB_t^v$ does not match with any trajectories, \emph{i.e.}, $\sum \vX(:,v)=0$. b) Trajectory $\calT_{t-1}^u$ is matched with detection $\calB_t^v$, \emph{i.e.}, $\vX(u,v)=1$. c)
	Trajectory $\calT_{t-1}^u$ does not match with any detections, \emph{i.e.}, $\sum \vX(u,:)=0$. In the following, we provide detailed descriptions of the trajectory management of different matching results.
	
	If $\sum \vX(:,v)=0$, \emph{i.e.}, the detection $\calB_t^v$ does not match with any existing trajectories. This indicates the detection is either a new occurred target or a false positive (FP) detection. Similar to \cite{wojke2017simple}, we initialize a ``tentative'' trajectory $\calT_t^i$ using $\calB_t^v$:
	\begin{equation}
	\calT_t^i = \{(\phi_{app}(\calB_t^v),\phi_{box}(\calB_t^v),\phi_{loc}(\calB_t^v))\},
	\end{equation}
	where $\phi_{app}(\cdot)$, $\phi_{box}(\cdot)$ and $\phi_{loc}(\cdot)$ output the appearance feature, bounding box and 3D location of the input detection, respectively. If the trajectory $\calT_t^i$ is matched with detections in coming image frames, the $\calT_t^i$ is then converted into a ``confirmed'' trajectory. Otherwise, we remove $\calT_t^i$ from trajectory list.
	
	If $\vX(u,v)=1$, the detection $\calB_t^v$ is assigned to trajectory $\calT_{t-1}^u$. We extend the trajectory $\calT_{t-1}^u$ using $\calB_t^v$:
	\begin{equation}
	\calT_t^u= \calT_{t-1}^u \cup \{(\phi_{app}(\calB_t^v),\phi_{box}(\calB_t^v),\phi_{loc}(\calB_t^v))\}.
	\end{equation}
	
	If $\sum \vX(u,:)=0$, \emph{i.e.}, the trajectory $\calT_{t-1}^u$ does not match with any detections, the target is temporally occluded or leaves the scene. If the trajectory is matched again in a following temporal window (such as within a coming 30 frames), we consider the target reappear after occlusion. Otherwise, we remove $\calT_{t-1}^u$ from trajectory list.
	
	\section{Experiment}
	\subsection{Dataset}
	We evaluate our proposed method on the JRDB dataset~\cite{martin2019jrdb}. This dataset contains over 60K data from 5 stereo cylindrical panorama RGB cameras and two Velodyne 16 LiDAR sensors. There are 54 sequences of 64 minutes captured from both indoor and outdoor environments, where 27 sequences are used for training and the others are for testing. The frame rate is at 15 FPS and the resolution is 752$\times$480. The dataset has over 2.3 million annotated 2D bounding boxes on 5 camera images and 1.8 million annotated 3D cuboids of over 3,500 targets.
	
	\subsection{Implementation Details}
	We follow the cascade detection paradigm~\cite{cai2018cascade} and adopt a ResNet50~\cite{he2016deep} as the backbone of our detector. We split the panorama image into 7 image slices with an overlap 0.2 along the image width, and the ground-truth annotations to different image slices accordingly. We augment the training data by mixup~\cite{zhang2017mixup} and multiscale augment. During training, the parameters of the detector are updated using an Adam optimizer~\cite{kingma2014adam} with a total number of 20 epochs, and the initial learning is set to $10^{-5}$. In the tracking, we adopt a ReID model~\cite{luo2019bag} as our feature extractor, which is pre-trained on the DukeMTMC dataset~\cite{ristani2016performance} using a triplet loss.

	\subsection{Evaluation Result} \label{subsec:eval_result}
	We compare the detection and tracking performance of our MMPAT on the JRDB dataset with state-of-the-art methods. For detection, we evaluate the proposed method in terms of Average Precision (AP $\uparrow$) and processing time (Runtime $\downarrow$). For tracking, we evaluate the MMPAT in terms of Multi-Object Tracking Accuracy (MOTA $\uparrow$) ~\cite{bernardin2008evaluating}, IDentity Switch (IDS $\downarrow$), False Positive (FP $\downarrow$), {and} False Negative (FN $\downarrow$). The $\uparrow$ indicates the higher is better, and $\downarrow$ is on the contrary.
	
	Table~\ref{tb:det} shows the detection results. We can see that, the proposed method significantly outperforms the other state-of-the-art method by a large margin (at least 15.7 improvement on AP) with a competitive processing speed (about 14 frames per second). This is a strong evidence that demonstrates the proposed detection algorithm is efficient for object detection in panorama image. In Table~\ref{tb:track}, compared with state-of-the-art method JRMOT, the proposed method significantly improves the tracking performance by a large margin (9.2 improvement on MOTA) by reducing IDS and FN number (reduce 25\% and 13\% IDS and FN, respectively),  while slightly worse on FP.
	
\begin{figure*}[tb]
		\centering
		\begin{tabular}{l}
			\includegraphics[width=0.85\linewidth,height=11cm]{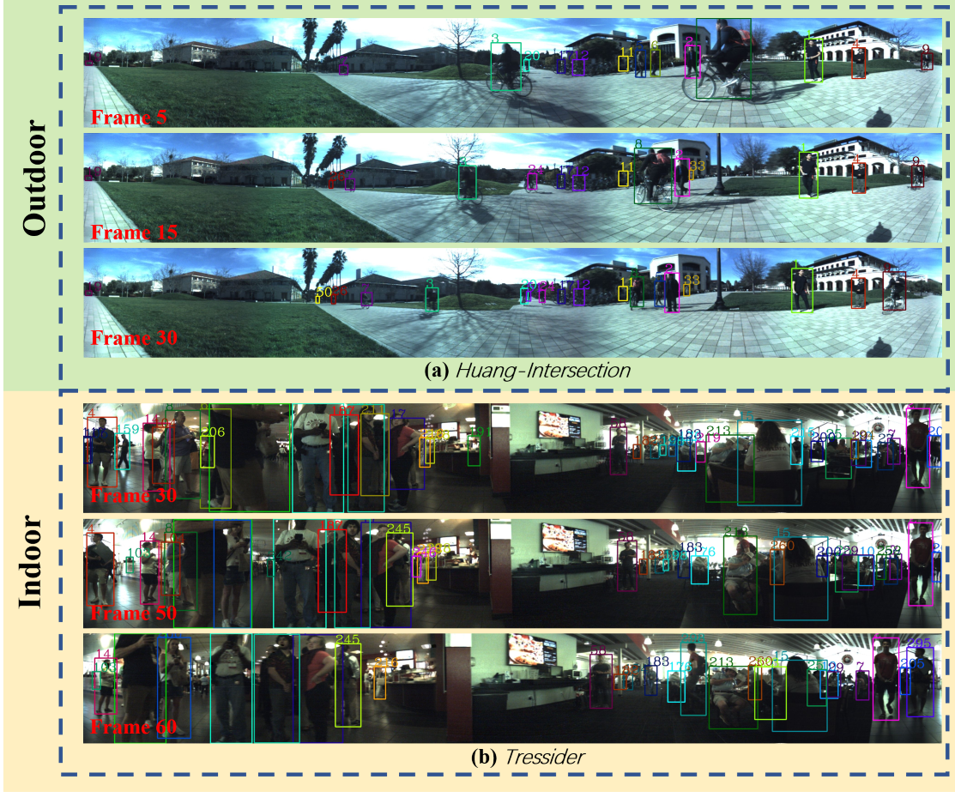}
		\end{tabular}
		\caption{Qualitative tracking results of (a) Huang-intersection (b) Tressider subsets of the JRDB dataset, where (a) is an outdoor scene and (b) is an indoor scene. The numbers upon bounding boxes denote ID labels.}\label{fig:track_result}
		\vspace{-0.3cm}
	\end{figure*}

	Figure~\ref{fig:track_result} illustrates some qualitative tracking results of the MMPAT on JRDB dataset. It can bee seen that, no matter in outdoor scenario with poor light conditions or in indoor scene with complex background clutters, the proposed method can robustly track targets and generate accurate trajectories for targets.
	
	\begin{table}[tb]
		\scriptsize
		\centering
		\caption{Detection results on the JRDB Dataset}\label{tb:det}
		\begin{tabular}{l|c|c}
			\hline
			Method 									& AP $\uparrow$ 	& Runtime $\downarrow$ \\
			\hline
			\hline
			YOLOV3~\cite{redmon2018yolov3}			& 41.73				& 0.051 \\
			DETR~\cite{carion2020end}				& 48.51				& 0.350  \\
			RetinaNet~\cite{lin2017focal} 			& 50.38				& 0.056 \\
			Faster R-CNN~\cite{ren2015faster} 		& 52.17				& \textbf{0.038} \\
			Ours				 					& \textbf{67.88}				& 0.070 \\
			\hline
		\end{tabular}
	\end{table}

	\begin{table}[tb]
		\scriptsize
		\centering
		\caption{Tracking Results On the JRDB Dataset}\label{tb:track}
		\begin{tabular}{l|c|c|c|c}
			\hline
			Method				& MOTA~$\uparrow$			& IDS~$\downarrow$ 	&  FP~$\downarrow$ & FN $\downarrow$ \\
			\hline
			\hline
			Tracktor~\cite{bergmann2019tracking}			& 19.7				& 7026		& 79573			  & 681672		\\
			DeepSORT~\cite{wojke2017simple}					& 23.2				& \textbf{5296}					& 78947			  & 650478		\\
			JRMOT~\cite{shenoi2020jrmot}					& 22.5				& 7719			& \textbf{65550}  & 667783		\\
			Ours											& \textbf{31.7}		& 5742				& 67171			  & \textbf{580565}	\\
			\hline
		\end{tabular}
		\vspace{-0.3cm}
	\end{table}
	
	\subsection{Ablation Study} \label{subsec:abla}
	In Table~\ref{tb:abla_det}, we provide ablation studies on the validation set of the JRDB dataset to analyze the influence of different components. The cascade r-cnn~\cite{cai2018cascade} is adopted as the \texttt{Baseline}. The \texttt{DCN} stands for deformable convolutional network~\cite{dai2017deformable}, \texttt{split} denotes splitting the panorama image into image slices, \texttt{mixup} denotes data mixing up, \texttt{multiscale} denotes multiscale testing, \texttt{softnms} denotes using the softnms. It can be seen that: 1) compared with the baseline method, our detector dramatically improves the detection performance by a large margin (17.9 improvement on AP). 2) The split of image can efficiently improves detection in panorama image (11.5 improvement on AP). This demonstrates that the response regions of targets in the feature maps can largely affect the detection performance. 3) Adding different data augmentation methods can steadily improves the performance.
	
	\begin{table}[tb]
		\scriptsize
		\centering
		\caption{Ablation Study on Object Detection}\label{tb:abla_det}
		\begin{tabular}{l|c}
			\hline
			Method		& AP $\uparrow$ \\
			\hline
			\hline
			Baseline										& 52.8			\\
			Baseline+DCN									& 53.1			\\
			Baseline+DCN+split								& 64.6			\\
			Baseline+DCN+split+mixup						& 68.2			\\
			Baseline+DCN+split+mixup+multiscale				& 69.7			\\
			Baseline+DCN+split+mixup+multiscale+softnms		& 70.7			\\
			\hline
		\end{tabular}
		\vspace{-0.3cm}
	\end{table}
	
	\section{Conclusion}
	This paper focuses on the multi-object tracking (MOT) problem of automatic driving and robot navigation. We propose a MultiModality PAnoramic multi-object Tracking framework (MMPAT), which takes both 2D 360$^\circ$ panorama images and 3D point clouds as input.  An object detection mechanism is designed to detect targets in panorama images. Besides, we also provide a 3D points collection algorithm to associate the point clouds with 2D images. We evaluate the proposed method on the JRDB dataset, which achieves the top performance in detection and tracking tasks and significantly outperforms state-of-the-art methods by a large margin (15.7 improvement on AP and 8.5 improvement on MOTA).
	
	\section{Acknowledge}
	This work is funded by National Key Research and Development Project of China under Grant No. 2019YFB1312000 and 2020AAA0105600, National Natural Science Foundation of China under Grant No. 62076195, 62006183, and 62006182, and by China Postdoctoral Science Foundation under Grant No. 2020M683489.
	
	{\small
		\bibliographystyle{ieee_fullname}
		\bibliography{egbib}
	}
	
\end{document}